\documentclass[runningheads]{llncs}
\usepackage{graphicx}

\usepackage{amsmath,amssymb,amsfonts}
\usepackage{algorithmic}
\usepackage{textcomp}
\usepackage{latexsym}
\usepackage{hyperref}
\usepackage{bm}

\usepackage[table]{xcolor}
\def\BibTeX{{\rm B\kern-.05em{\sc i\kern-.025em b}\kern-.08em
    T\kern-.1667em\lower.7ex\hbox{E}\kern-.125emX}}

\begin{document}
\title{Confidence Preservation Property in Knowledge Distillation Abstractions}
%
%
\author{Dmitry Vengertsev\inst{1}\orcidID{0000-0002-6039-0579} \and
Elena Sherman \inst{1}\orcidID{0000-0003-4522-9725} }
\authorrunning{D. Vengertsev et al.}
%
\institute{Department of Computer Science, Boise State University, Boise, ID 83725, USA \\ \email{	dmitryvengertsev@u.boisestate.edu, elenasherman@boisestate.edu} }
\maketitle              
\begin{abstract}
Social media platforms prevent malicious activities by detecting harmful content of posts and comments. To that end, they employ large-scale deep neural network language models for sentiment analysis and content understanding. Some models, like BERT, are complex, and have numerous parameters, which makes them expensive to operate and maintain. To overcome these deficiencies, 
industry experts employ a knowledge distillation compression technique, where a distilled model is trained to reproduce the classification behavior of the original model.

The distillation processes terminates when the distillation loss function reaches the stopping criteria. This function is mainly designed to ensure that the original and the distilled models exhibit alike classification behaviors. However, besides classification accuracy, there are additional properties of the original model that the distilled model should preserve to be considered as an appropriate abstraction.

In this work, we explore whether distilled TinyBERT models preserve confidence values of the original BERT models, and investigate how this confidence preservation property could guide tuning hyperparameters of the distillation process.

\end{abstract}

\section{Introduction} \label{sec:Intro}



Deep neural language models such as BERT~\cite{devlin2018bert}, ELMo~\cite{elmo2018}, GPT-3~\cite{brown2020language} play a crucial role in screening social media for fake and harmful content~\cite{heidari2020using}. To reliably classify information, those models require high precision, which often comes at the cost of increased the model's size~\cite{kaplan2020scaling}. However, larger size models have high inference latency and are problematic to deploy on mobile devices and embedded systems. A traditional approach by verification community is to reduce large systems while preserving desirable properties through abstraction~\cite{Loiseaux:FMSD95}. There are several examples of constructing explicit abstraction: conversion of neural network to Boolean combinations of linear arithmetic constraints~\cite{pulina2010abstraction}, empirical extraction of a deterministic finite automaton using clustering of the hidden states of deep neural networks~\cite{wang2018verification}, and leveraged pruning during verification process~\cite{guidotti2020verification}. An orthogonal approach to creating smaller models through the knowledge transfer (training) from a larger model without a significant drop in accuracy is known as knowledge distillation. However, the majority of current distillation techniques focuses on the classification correctness while ignoring other properties of the teacher model.

 In this work, we investigate whether the knowledge distillation~\cite{hinton2015distilling} used in TinyBERT behaves as an abstraction of a large BERT model and preserves pairwise confidence property. We call this type of abstraction \emph{implicit} since unlike the previously mentioned clustering and pruning abstraction techniques, it has no direct mappings between BERT and TinyBERT internal architectures. 
 

 Our proposed approach determines whether a distilled model preserves properties of the original model through black-box equivalence checking~\cite{Dahiya:APLAS17}. In particular, we investigate preservation of the confidence property, which is important for high-stake real-world applications such as AI-aided medicine and pedestrian detection. In the first application, the control should be given to a doctor when the confidence of the diagnostics model is low \cite{esteva2017dermatologist}, and in the second application when pedestrian detection network is not able to confidently predict the presence or absence of immediate obstructions, the car should rely more on the output of other sensors for braking.

In this paper, we outline the concept of confidence preservation property and establish a method for assessing it. 
To uncover the confidence-related distillation disagreements between the teacher and the student models, we introduce a pairwise confidence difference, Fig.~\ref{fig:sigma_plots}. We evaluate the preservation of this property on six tasks from the  General Language Understanding Evaluation (GLUE) dataset. Our results show that the distilled TinyBERT model maintains the confidence property of the original BERT model in three tasks. For the remaining three tasks, whose models fail to preserve the property, we modify the training hyperparameters so that these models maintain the confidence property without degrading the original accuracy. 


Overall, this work has the following contributions: (1) Considering knowledge distillation as an implicit abstraction with anticipated property preservation characteristics. (2) A confidence property preservation criterion based on pair-wise confidence measurements and its empirical evaluations. (3) Identifying and empirically tuning hyperparameters in the distillation process to ensure the confidence property preservation.

\section{Background and Motivation} \label{sec:background}

 \begin{figure*}[t] 
  \centering \includegraphics[scale=0.6]{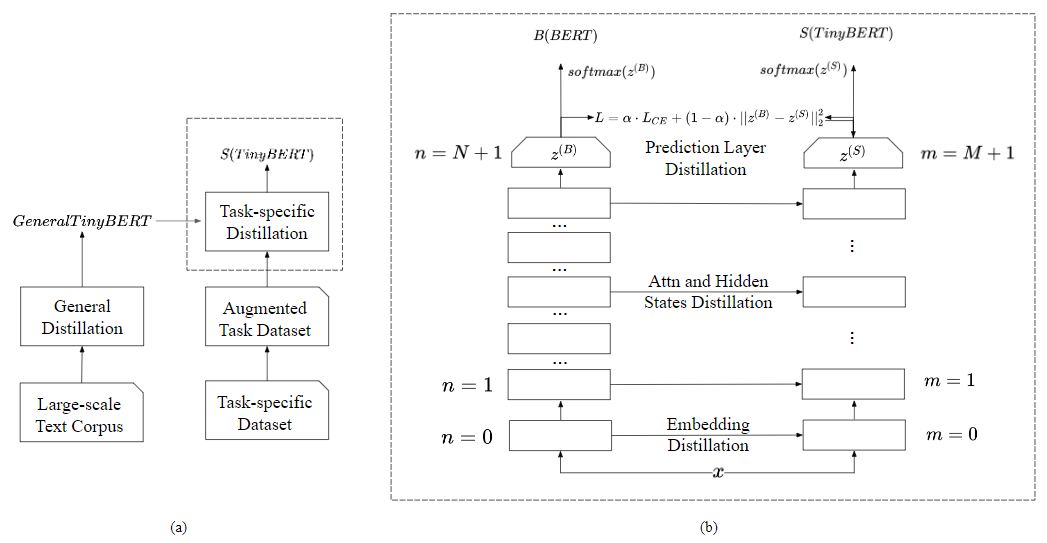}
  \caption{Learning abstract model via distillation. (a) end to end distillation flow; (b) task specific distillation}
  \label{fig:distillation}
\end{figure*}

\subsection{Significance of Knowledge Distillation Models}
Recent advancements in the area of machine learning are driven by transformer-based models.
Unfortunately, the size and the efficiency of these models prohibit their use in resource-constrained and time-sensitive applications. For example, a baseline transformer model executes a single query in a few tenths of seconds, which is slow for real time systems. To reduce desirable response latency to milliseconds~\cite{kim2020fastformers} compressed models that are obtained via knowledge distillation technique~\cite{bucilua2006model,hinton2015distilling} are used. The technique essentially trains a compact model, referred to as the student, to reproduce the behavior of a larger model, known as the teacher.

In this paper we focus on distillation of BERT (Bidirectional Encoder Representations from Transformers) model~\cite{vaswani2017attention}, which has significantly advanced the state-of-the-art in natural language processing tasks such as language understanding, text classification, and language translation. 
Previous work distills BERT into smaller models such as single layer BiLSTM~\cite{tang2019distilling}, DistilBERT~\cite{sanh2019distilbert}, 
and TinyBERT~\cite{jiao2020tinybert}. 
In this paper, we examine property preservation of TinyBERT models since they outperform other distillation methods and also combine both response and feature-based knowledge transfers. The reason TinyBERT achieves superior accuracy is due to its pre-training feature and a specialized loss function. At the pre-training stage, obtaining a good initialization is crucial for the distillation of transformer-based models. Intermediate loss during feature-based distillation provides a boost of classification accuracy of at least 10\% for most of the GLUE tasks~\cite{jiao2020tinybert}.

\subsection{TinyBERT Distillation}

First, we explain the TinyBERT distillation process using the diagrams shown in Fig.~\ref{fig:distillation}, and then discuss the distillation loss function.

At the general distillation stage (Fig.~\ref{fig:distillation}.a), the original BERT model acts as the teacher $B$. The student model GeneralTinyBERT mimics the teacher’s behavior on general-domain corpus from Wikipedia by using the feature-based distillation, rather than response-based distillation. General distillation helps TinyBERT learn the rich knowledge embedded in pre-trained BERT, which plays a major role in improving the generalization capability of the task-specific TinyBERT.

The general distillation produces GeneralTinyBERT that is used as the initialization of the student model $S$ for the task-specific distillation. This stage first completes the data augmentation, and then performs distillation on the task-specific augmented dataset.
As shown in the Fig.~\ref{fig:distillation}.b, both the teacher $B$ and the student $S$ models have the same type of the layers, however student model has $m=1,..,M$ transformer layers and teacher model has $n=1,..,N>M$ such layers. Embedding layers for the student and the teacher are $m=0$ and $n=0$ correspondingly, and the prediction layers are labelled $m=M+1$ and $n=N+1$.

An input text sequence $x$ enters the embedding layer, and then transformer layers. A single transformer layer includes the two main sub-layers: multi-head attention (MHA) and fully connected feed-forward network (FFN). MHA is described by attention matrices, and FFN is described using hidden states. The output of the last transformer is passed to the prediction layer that outputs raw predictions in the form of logits $z$. Logits go through the softmax layer for the final task-specific classification.

There are several ways to map the $N$ layers of teacher model to the $M$ layers of the student model. We use the uniform layer mapping strategy as it provides a superior accuracy for the majority of the tasks \cite{jiao2020tinybert}. This strategy covers the knowledge transfer from bottom to top layers of BERT to the corresponding layers of the student model. During the training of a student model, TinyBERT uses four loss functions: embedding distillation loss for $m=0$, attention and hidden distillation losses for transformer layers $0\leq m\leq M$, and finally, the distillation of the prediction layer. This work focuses on the distillation of the prediction layer, rather than intermediate layers.

At the prediction layer the loss is defined in Eq.~\ref{eq:loss_prediction_layer}, where the student cross entropy loss $L_{CE}$ is supplemented with the distillation objective $L_{dist}$ in order to penalize the loss between the student network’s logits against the teacher’s logits, which determines distillation training objectives:

 \begin{equation} \label{eq:loss_prediction_layer}
 \begin{split}
    L & = \alpha \cdot L_{CE} + (1 - \alpha)\cdot L_{distill} \\
  &   = - \alpha \sum_{i}t_i\log y_i^{(S)} +  (1-\alpha)\sum_{x\in X^{train}}\lVert z^{(B(x))}-z^{(S(x))}\rVert^2
 \end{split}
\end{equation}
where $z^{(B)}$ and $z^{(S)}$ are the logits of the teacher $B$ and the student $S$, correspondingly. The variable $t$ is the one-hot target ground truth label, $\alpha \in [0,1]$ is a constant, $i$ is the index of the training examples from the augmented dataset.

\subsection{Distillation as Implicit Abstraction}

Abstraction is a powerful technique that is used extensively in software verification. The key to an abstraction is to establish a relationship between a concrete system $P$ and its model $M$ that hides unnecessary details about $P$ while preserving some essential $P$'s properties~\cite{mine2002few,clarke2000counterexample,Loiseaux:FMSD95}. Because the relationship is established, we refer to this type of abstraction as an explicit abstraction.

Previous research performs such explicit abstraction on a deep neural network using a clustering approach~\cite{wang2018verification}, where a set of nodes in the neural network relates to a single node in the abstracted model. In another approach, researchers eliminate less relevant nodes in a neural network using pruning process~\cite{guidotti2020verification}. All these techniques indeed reduce the size of the original model to make it more amenable, for example, for verification. However, the drawback of these techniques is that the accuracy of the resulting abstraction suffers, i.e., they fail to preserve the classification accuracy of the original neural network.
This is because they do not consider accuracy preservation in their abstraction process. Clearly, considering the size of neural networks, identifying such explicit abstraction is a daunting task.

On the contrary, the distillation process ensures that the student model has similar accuracy as the teacher model. Since, by construction, the distilled model is smaller in terms of layers and dimensions, we can essentially consider it as an abstraction of the original model. However, since the student model might have a completely different architecture than its teacher, establishing an explicit mapping between those two systems is challenging. Instead, we say that these models have an \textit{implicit} abstraction, which is created based on correctness preservation. Thus, the question is whether this implicit abstraction preserves properties between two models. One of the properties that we investigate here is confidence, which we discuss in the next section.

\section{Confidence Property Preservation Criterion} \label{sec:properties}





In the real-world classification problems, in addition to a model's accuracy, the level of confidence with which a model performs classification, (i.e., predictive probability) is also considered. Confidence value can be used to determine the model's properties, such as high-confidence and decisiveness~\cite{vengertsev2020recurrent}. Therefore, preserving these values of the original model should be an important feature of the distilled abstraction.


The literature offers different measurements of the confidence for neural networks. Traditionally, it is defined as the maximum element of the probability vector obtained at the final softmax layer of a model $M$ for an input $x$, i.e.,
\begin{equation}
\label{eq:conf}
\mathsf{Cnf}^M(x) = max(\mathsf{softmax}(z^{M(x)}))
\end{equation}

Recent work has shown that using this definition might be inadequate~\cite{pearce2021understanding} and general purpose method based on Bayesian interpretation of the neural networks~\cite{gal2016uncertainty}, or confidence calibration~\cite{guo2017calibration} should be used. Beyond being uncalibrated, the main criticisms of the softmax-based confidence evaluation is its failure to decrease confidence value on inputs far from the training data~\cite{pearce2021understanding,szegedy2013intriguing}. Therefore, this work evaluates confidence preservation on the training data set $X^{train}$, and we monitor ECE of both student and teacher models, as well as accuracy and the pairwise confidence on the evaluation dataset.



\setlength{\abovecaptionskip}{5pt plus 2pt minus 1pt} 
\begin{figure}[t]
  \centering \includegraphics[scale=0.5]{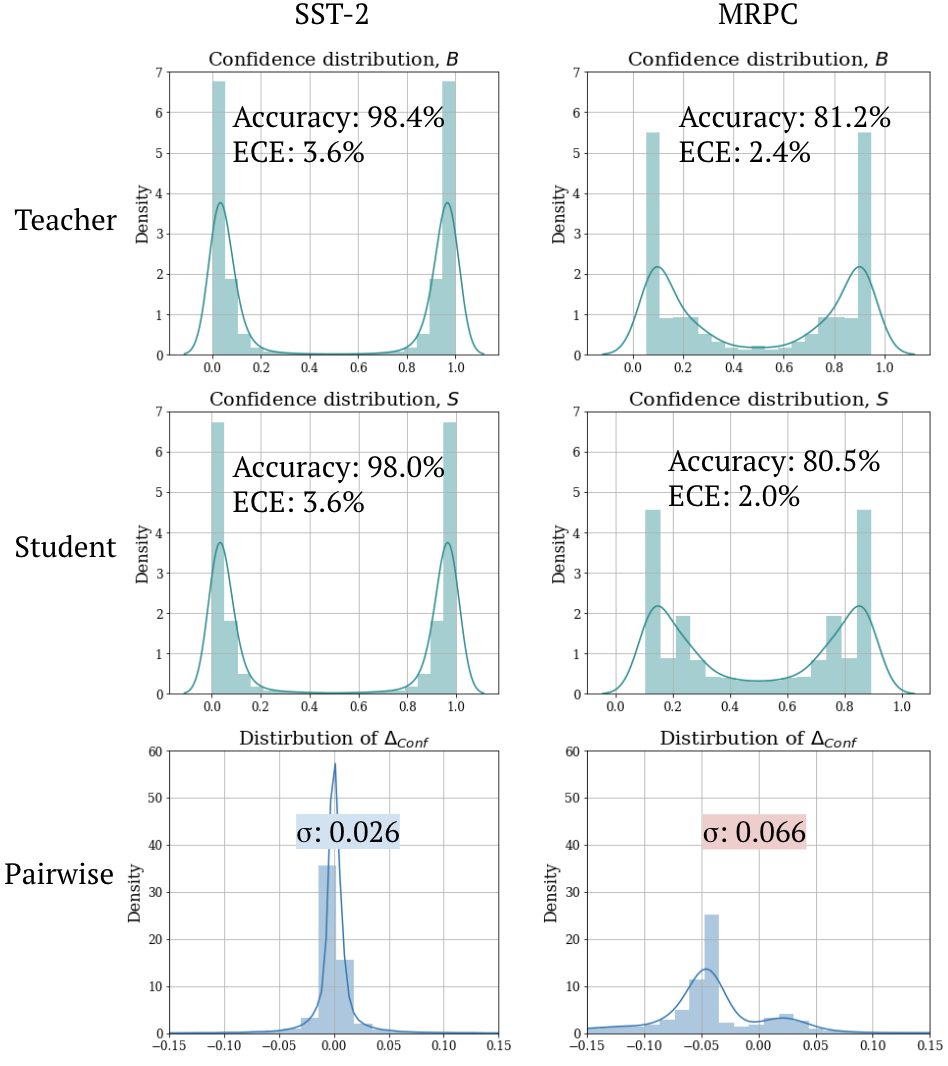}
  \caption{For two linguistic tasks: SST-2 and MRPC, the individual distributions of softmax confidence for the teacher and the student do not show significant difference, under comparable expected calibration error (ECE). However, the distribution of the pairwise confidence does highlight the issue of poor distillation for the MRPC task.}
  \label{fig:sigma_plots}
\end{figure}

\subsection{Pairwise Confidence Preservation Property} 
In order to determine how to define and measure the confidence preservation property, we first examine confidence values for both $S$ and $B$ models. The confidence values $\mathsf{Cnf}(x)$ of a model $M$ over $X^{train}$ represent a distribution in range $[0,1]$. To better understand the data, we first plotted confidence values for $B$ and $S$ models as density plots. Fig.~\ref{fig:sigma_plots} shows these distribution for SST-2 and MRPC benchmarks: in the first top row for $B$ and in the second row for $S$.

From these plots, we can see that the traditional metric of aggregated confidence distributions of $S$ and $B$ models, as well as expected calibrated error~(ECE)~\cite{guo2017calibration} are not good candidates for identifying differences in confidence. Two plots might have points with the same confidence value for different inputs, which is misleading. That is, such aggregate data loses information on whether confidence is preserved on the same inputs. Likewise, the ECE values of the models can be similar as in the case of MRPC, e.g., 2.4 vs. 2.0, but actually fail to hold the confidence preservation property. 

Thus, our main idea is to measure the confidence preservation on the same input example for two models $B$ and $S$. In other words, in this work we focus on confidence preservation in the context of \textbf{functional equivalence} where for the same input $B$ and $S$ should have similar confidence values. The notion of functional equivalence is commonly used for verification of traditional programs~\cite{huang2012formal}. Thus, we adapt a similar idea of ``input-output'' equivalence to define the \textbf{property of confidence preservation} for the case of deep neural networks.

Therefore, to measure this property, we introduce a pairwise confidence difference to uncover the confidence-related distillation disagreements between $B$ and $S$ models. The last row in Figure~\ref{fig:sigma_plots} depicts the pair-wise difference distribution and their difference spread $\sigma$. We use $\sigma$'s value and its threshold of $0.05$ as the property preservation criterion. As the data in the graphs show in this example the SST-2 model satisfies the property preservation criterion, while the MRPC fails do so since its $\sigma > 0.05$.



Here we formally describe the measurement of the differences between $\mathsf{Cnf}^B(x)$ and $\mathsf{Cnf}^S(x)$. First, let's define a function $\mathsf{Idx}^M(x)$ that returns an index in the probability vector of the maximum element return by Eq.~\ref{eq:conf}. Then we define the pairwise confidence difference for $\forall x \in X^{train}$ as follows:
\begin{equation}
\Delta_{Cnf}(x) = 
   \begin{cases}
     \mathsf{Cnf}^S(x) - \mathsf{Cnf}^B(x) & \text{if } \mathsf{Idx}^B(x) = \mathsf{Idx}^S(x) \\
     \bot & \text{otherwise}
   \end{cases}
\end{equation}





Since $\Delta_{Cnf}$ can have both positive and negative values, we use the sum of squares to compute the effect of these differences. Thus, we propose the following formula to compute the confidence difference on a training set $X^{train}$
$
 \sigma(X^{train}) = \sqrt{\frac{1}{|X^{train}|}\sum_{x\in X^{train}}\Delta^2_{Cnf}(x)}
$

\begin{definition}{\bf Pairwise Confidence Preservation Property $\bm{\varphi_{cnf}}$} 
We say that the confidence preservation property $\bm{\varphi_{cnf}}$ holds if for any input $x \in X^{train}$ the spread of the pairwise confidence differences $\sigma$ between distilled $S$ and original $B$ models is small on $X^{train}$:
\begin{equation} \label{eq:confidence_property}
 \sigma(X^{train}) \leq \kappa
\end{equation}
where $\kappa>0$ is a small constant, which is determined by users. 
\end{definition}

\subsection{Confidence Preservation Property $\bm{\varphi_{cnf}}$ Dependencies}
Since a traditional distillation process does not take into  account the confidence property preservation as in Definition~(\ref{eq:confidence_property}), the technique does not guarantee it to hold for all models. In this section, we establish dependencies between $\sigma(X^{train})$ and the distillation parameters that demonstrate the existence of parameters that can be tuned for the property~(\ref{eq:confidence_property}) to hold.






 To determine such relations, we first define a \textbf{distillation quality condition}  on student loss $L^{(S)}$ defined in (\ref{eq:loss_prediction_layer})
 \begin{equation} \label{eq:distil_condition}
     L^{(S)}\leq\beta
 \end{equation}
 where $\beta$ is a small positive scalar.
 



Next, from the softmax property \cite{gao2017properties} (Proposition 4) we obtain the following inequality for any input $x$: $
         \| \mathsf{softmax}(z^{S(x)}) - \mathsf{softmax}(z^{B(x)}) \| < \gamma \| z^{S(x)} -  z^{B(x)}\|  $
, where $\gamma$ is the inverse temperature constant in the softmax function. 

Since the norm of a vector is always greater than or equal to any individual value of the vector, we have:
\begin{equation} \label{eq:vector_vs_norm}
       \Delta_{Cnf}(x) \leq  \| \mathsf{softmax}(z^{S(x)}) - \mathsf{softmax}(z^{B(x)}) \| 
\end{equation} 
Therefore, from (\ref{eq:vector_vs_norm}) we obtain:
        \begin{equation} \label{eq:softmax_ineq}
         \sum_{x\in X^{train}}\Delta_{Cnf}^2(x) < \gamma^2 \sum_{x\in X^{train}}\| z^{S(x)} -  z^{B(x)}\|^2 
        \end{equation}

        \begin{equation} \label{eq:test}
        \sigma(X^{train})^2|X^{train}| =\sum_{x\in X^{train}}\Delta_{Cnf}^2(x) < \gamma^2 \sum_{x\in X^{train}}\| z^{S(x)} -  z^{B(x)}\|^2  < \gamma^2\frac{\beta}{(1-\alpha)}
        \end{equation}

       \begin{equation} 
       L_{dist} \leq \frac{\beta}{1-\alpha}
       \end{equation}

Recall in Eq.~\ref{eq:loss_prediction_layer} the student distillation objective $L^{(S)}$ consists of two positive addends: student's cross entropy loss $L_{CE}$ and the distillation objective $L_{dist}$. If $\alpha L_{CE} + (1-\alpha) L_{dist} \leq \beta$, then $L_{dist} \leq \frac{\beta}{1-\alpha}$ holds because the value of $L_{CE}$ is positive as the logarithm of a value that is between zero and one is negative. Therefore, from the distillation quality condition (\ref{eq:distil_condition}) we obtain:
    $\sum_{x\in X^{train}}\lVert z^{(S(x))}-z^{(B(x))}\rVert^2 < \frac{\beta}{(1-\alpha)}$
    
Finally, using Eq.~\ref{eq:softmax_ineq} we produce:
    $\sigma(X^{train}) < \gamma \sqrt{\frac{\beta}{|X^{train}|(1-\alpha)}}$
, which means (\ref{eq:confidence_property}) is satisfied with $\kappa=\gamma \sqrt{\frac{\beta}{|X^{train}|(1-\alpha)}}$. This means that the confidence preservation property depends on several parameters. 

However, some of those parameters are predefined for TinyBERT models. Thus, in the TinyBERT model a regular softmax function is used, which sets the inverse temperature constant $\gamma=1$. Moreover, due to the multi-stage approach in TinyBERT, during the prediction layer distillation only $L_{distill}$ is used, so $\alpha=0$. As a result, the confidence property preservation mainly depends on the square root of the distillation quality condition $\beta$ and the training set size. That is 
    \begin{equation} \label{eq:sigma_upper_bound}
\sigma(X^{train}) < \sqrt{\frac{\beta}{|X^{train}|}}
    \end{equation}

We assume that the training set is fixed, due to the cost of the data collection, and the main focus is on $\beta$ that depends on the distillation training hyperparameters. We focus on empirical search and fine-tuning to obtain $S$ models for which $\varphi_{cnf}$ holds.

\section{Experiment Setup} \label{sec:experiment_setup}


To investigate preservation $\varphi_{cnf}$ as defined in Eq.~\ref{eq:confidence_property} for TinyBERT models and answer our research questions, we use the training benchmarks for five language tasks with different TinyBERT settings. In this section, we describe datasets, knowledge distillation model settings, and the choice of the threshold parameter $\kappa$. In the next section, we state our two research questions and present evaluation results that answer them.

\subsection{GLUE Tasks Benchmarks}
The datasets used for training and evaluation during the knowledge distillation consist of the six benchmarks from the General Language Understanding Evaluation (GLUE)~\cite{wang2018glue}

The selected benchmark tasks are sentiment similarity and the natural language inference for binary classification tasks. 
Table~\ref{table:datasets} summarizes each dataset/task, with the information on both training (Tr) and evaluation (Ev) dataset sizes. For the smaller datasets such as CoLA, RTE, and MRPC we perform data augmentation. Augmentation is only performed for training datasets, we still perform evaluation on non-augmented data. 

For the data augmentation we use the same approach as in work by Jiao et. al~\cite{jiao2020tinybert} relying on GLOVE dataset~\cite{pennington2014glove}. Specifically, we adopt \texttt{glove.6B.300d} that maps text sequences to a 300 dimensional dense vector space. 
As a result of data augmentation the sizes of the small training datasets - CoLA, RTE and MRPC - have increased to 211,457, 189,277 and 224,132, respectively; the original sizes of these datasets are shown in the parentheses in the table.



\begin{table}
	\caption{Benchmark datasets GLUE \cite{wang2018glue} }
	\centering 
	\label{table:datasets}
	\begin{tabular}{l l l r } 
		\hline 
		Task & Description & Tr & Ev  \\ [0.5ex]
		\hline 
		SST-2 & Sentiment analysis & 67,349  & 872  \\ 
  	RTE & Natural language inference & 189,277 (2,490) & 277  \\ 
		QQP  & Paraphrase similarity & 363,846  & 40,430  \\ 
		QNLI & Natural language inference & 104,743 & 5,463  \\
		MRPC & Semantic textual similarity & 224,132 (3,668) & 408   \\ 
  		CoLA & Semantic correctness & 211,457 (8,551) & 1,043   \\ 
		\hline 
	\end{tabular}
	\label{table:lstm_transition} 
\end{table}

\subsection{Model Settings}
We use a task-specific fine-tuned BERT model $B$ with $N=12$ transformer layers. Specifically, we use an uncased BERT version, meaning the input text has been changed to lowercase at the input layer. The model $B$ has the dimension of the hidden layer $d^{'}$=768, with the total number of model parameters at 109 million. 

For the students, we use two models $S_{4L}$ and $S_{6L}$. $S_{4L}$ is a smaller distilled model with only four transformer layers $M$=4 and relatively small dimension of a hidden layer $d^{'}$=312. In total $S_{4L}$ has 14.5M model parameters. For $S_{6L}$ $M$=6, $d^{'}$=768, resulting in 67M model parameters.

For general distillation, we use the model by the authors of TinyBERT~\cite{jiao2020tinybert}, which is distilled from BERT using domain agnostic corpus from English Wikipedia (2,500M words), sequence length of 128 and feature-based distillation. We do not alter this general distillation model in this work, since the focus is on the task-specific models and prediction-layer distillation.

We perform task-specific distillation for the above six tasks, which produces twelve TinyBERT models. We initialize the student model with the parameters of GeneralTinyBERT, and for the teachers we use BERT that is fine-tuned for the corresponding task. We use the input sequence length of 128 for task specific distillation. As for the learning parameters, we use learning rate of $3e^{-5}$, batch size of $32$, the number of epochs for intermediate distillation (embedding layer, attention matrices and hidden layer) is $10$ and the number of epochs for prediction layer distillation is $3$. The resulting accuracy numbers of the TinyBERT that we distilled from the BERT are comparable to the ones presented in the original paper~\cite{jiao2020tinybert}, thus making our reproduction of TinyBERT valid. 

We perform the knowledge distillation on GPU NVIDIA V100 computation with 16 GB RAM running on top of Google Cloud Platform (GCP) service. 



\subsection{Parameters Selection} 

As described in Section~\ref{sec:properties}, the confidence preservation property $\varphi_{cnf}$ is parameterized by the threshold $\kappa$ in Eq.~(\ref{eq:confidence_property}). We select this threshold as $\kappa$=0.05 for our adequacy criterion for $\varphi_{cnf}$ to hold. 
The final condition we evaluate is:
\begin{equation*}
    \sigma(X^{train})\leq 0.05
\end{equation*}
That is, we say that \textbf{confidence preservation holds} if the value of $\sigma$ on the training set $X^{train}$ is less or equal to the threshold $0.05$.


To avoid negative effect of hyperparameter fine-tuning on $S$'s accuracy, we add a constraint on changes in accuracy value of $S$ that prevents \textit{significant accuracy drop}. We consider a drop of accuracy below $1\%$ to be significant, as according to a Jiao et al.~\cite{jiao2020tinybert} it corresponds to the loss of the ``on-par'' performance. Thus, the accuracy of the fine-tuned $\Tilde{S}$ and the original $S$ cannot be less than 1\%.

\section{Experimental Evaluations and Results}  \label{sec:experiments_result}
We conduct our empirical evaluations to answer the following research questions:

\begin{itemize}
 \item  \textbf{RQ1: Confidence preservation prevalence}. Do the distilled models $S$ from $B$ preserve $\varphi_{cnf}$ property?

\item \textbf{RQ2: Confidence preservation dependencies}. Can tuning the distillation hyperparameters of a failed model $S$ make the property $\varphi_{cnf}$ hold for its tuned model $\Tilde{S}$? 



  
  
\end{itemize}

\subsection{Confidence Preservation Prevalence (RQ1)} 

In this section, we evaluate whether $\varphi_{cnf}$ holds for the six language tasks. In the Table-\ref{table:rq1_results}, 
%
 for each task and the dataset we have two multi-column headers ``Models'' and ``$\varphi_{cnf}$ property''. The former describes individual model performance metrics such as accuracy (Acc) and expected calibration error (ECE) for three models $B$, $S_{4L}$, and $S_{6L}$. The $\varphi_{cnf}$ property columns contain data pertaining to evaluations of $\varphi_{cnf}$. Each row corresponds to a task dataset evaluated on the training (Tr) and evaluation (Ev) sets.
%
As we discussed in the beginning of Section~\ref{sec:properties}, we examine $\varphi_{cnf}$ on Tr dataset, due to the need to remove the factor that can affect confidence correctness on the inputs that are far from the training dataset.
We do present results for Ev dataset as well to demonstrate the consistency of our experiments, i.e., if $\varphi_{cnf}$ does not hold on Tr, then we expect it to perform the same way on Ev, as well as to observe the same trend in the confidence value change for $\Tilde{S}$ models. 

\begin{table*}[t]
	\caption{Confidence Preservation Property $\varphi_{cnf}$ for GLUE language tasks}
	\centering 
	\label{table:rq1_results}
	\begin{tabular}{l |  c| c c| c c| c c |c c} 
\hline 
         &   &  \multicolumn{6}{c}{Models} & \multicolumn{2}{|c}{$\varphi_{cnf}$ property}  \\
         \cline{3-10}
         Task & Dataset &  \multicolumn{2}{c|}{$B$} &  \multicolumn{2}{c|}{$S_{6L}$}  &  \multicolumn{2}{c|}{$S_{4L}$} & $B$ vs $S_{6L}$ & $B$ vs $S_{4L}$  \\
           &  &  Acc & ECC & Acc & ECC &  Acc & ECC  & $\sigma(X) $ & $\sigma(X)$ \\
		\hline 

        SST-2 &  Tr &  98.4 & 3.6 & 98.0 & 3.6 & 97.0 & 3.5  & \cellcolor{gray!15}0.026 & 0.055 \\
         &  Ev &  93.0 & 1.6 & 91.4 & 1.3 & 89.4 & 2.9  & 0.046 & 0.075 \\
         \hline 
        RTE &  Tr & 94.8 & 22.1 & 84.3 & 14.9 & 86.5 & 21  & \cellcolor{gray!40}0.062 & 0.109 \\
         &  Ev &  65.3 & 9.0 & 66.8 & 9.9 & 60.6 & 2.5  & 0.100 & 0.107 \\
         \hline 
        QQP &  Tr &  96.9 & 7.4 & 95.2 & 7.5 & 92.2 & 6.0  & \cellcolor{gray!15}0.049 & 0.077  \\
         &  Ev &  90.7 & 2.7 & 90.9 & 4.2 & 89.1 & 3.5  & 0.055 & 0.079  \\
         \hline 
        QNLI &  Tr &  97.6 & 9.3 & 96.0 & 9.3 & 91.3 & 4.8  & \cellcolor{gray!15}0.049 & 0.065\\
         &  Ev &  91.4 & 4.8 & 91.1 & 5.8 & 85.8 & 1.8   & 0.079 & 0.094\\
         \hline 
        MRPC &  Tr &  81.2 & 2.4 & 80.5 & 2.0 & 79.9 & 3.7   & \cellcolor{gray!40}0.066 & 0.054 \\
         &  Ev &  78.7 & 7.5 & 75.5 & 2.3 & 74.2 & 7.2  & 0.070 & 0.075 \\
        \hline 
        CoLA &  Tr &  98.5 & 5.3 & 96.6 & 5.7 & 94.7 & 6.4  & \cellcolor{gray!40}0.059 & 0.083 \\
         &  Ev &  83.2 & 7.6 & 79.6 & 7.4 & 72.6 & 11.5  & 0.098 & 0.129 \\
        
		\hline 
	\end{tabular}
\end{table*}

The first observation from the table is that for all language tasks, none of $S_{4L}$ models maintain $\varphi_{cnf}$. As a result, we conjecture that distilling BERT to just four transformer layers with the smaller dimension of hidden states (resulting in a 7.5X reduction in parameters) is too aggressive to satisfy $\varphi_{cnf}$.

The second and more important result is that the $S_{6L}$ model satisfies $\varphi_{cnf}$ for the three language tasks (shaded in \colorbox{gray!15}{light gray}), however it fails to do so for the three other tasks (shaded in \colorbox{gray!40}{dark gray}).

The answer to RQ1 is that the knowledge distillation does not uniformly preserve $\varphi_{cnf}$ across all six linguistic tasks. This conclusion supports our intuition that a standard distillation process focuses on classification accuracy and does not take into account confidence of predicted classes. Therefore, in the next research question RQ2, we investigate how parameters tuning can help with the tasks for which $\varphi_{cnf}$ fails.

\begin{table*}[t] 	
	\caption{Improving $\varphi_{cnf}$ property}
	\centering 
        \label{table:rq2_results}
        \def\arraystretch{1.2}%
	\begin{tabular}{l |  c| c | c   c  |c c  |c c |c c } 
    \hline 
         &   &  \multicolumn{5}{c}{Models} & \multicolumn{4}{|c}{$\varphi_{cnf}$ property}  \\
         \cline{3-11}
         Task & Dataset &  \multicolumn{1}{c|}{$B$} &  \multicolumn{1}{c}{$S_{6L}$} &  \multicolumn{1}{c|}{$\Tilde{S}_{6L}$}  &  \multicolumn{1}{c}{$S_{4L}$} & \multicolumn{1}{c|}{$\Tilde{S}_{4L}$} & $B$ vs $S_{6L}$ & $B$ vs $\Tilde{S}_{6L}$ & $B$ vs $S_{4L}$  & $B$ vs $\Tilde{S}_{4L}$  \\
           &  &  Acc &  Acc &  Acc & Acc & Acc & $\sigma(X)$ & $\sigma(X)$  & $\sigma(X)$ & $\sigma(X)$ \\
		\hline 
        RTE &  Tr & 94.8 & 84.3  & 84.6 &  86.5   & 88.5 &  \cellcolor{gray!40}0.062 & \cellcolor{gray!15}0.050& 0.109 & 0.069\\
         &  Ev &  65.3 & 66.8 &  66.4 &  60.6 & 62.5 & 0.100 & 0.084 & 0.107 & 0.119\\
         \hline 
        MRPC &  Tr &  81.2 &  80.5 & 80.2 & 79.9 & 79.4 & \cellcolor{gray!40}0.066 & \cellcolor{gray!15}0.047 &  0.054 & 0.053\\
         &  Ev &  78.7 &  75.5 &  75.9 &  74.2 &  73.7 & 0.070 & 0.060 & 0.075 & 0.074 \\
         \hline 
        CoLA &  Tr &  98.5 &  96.6 & 98.1 & 94.7 &  95.8 &  \cellcolor{gray!40}0.059 & \cellcolor{gray!15}0.039 & 0.083 & 0.071\\
         &  Ev &  83.2 & 79.6  & 80.8  & 72.6 & 72.1 & 0.098 & 0.093 & 0.129 & 0.120 \\        
		\hline 
	\end{tabular}
\end{table*}

\subsection{Confidence Preservation Dependencies (RQ2)} 

We considered several hyperparameters that can improve $\varphi_{cnf}$ adequacy for failed models $S$. In particular, we investigated the tasks that have inadequate values of $\sigma(X^{train})$ that fail $\varphi_{cnf}$ - RTE, MRPC, and CoLA with the original recommended TinyBERT parameters: batch size 32, three epochs, weight decay $1e^{-4}$, intermediate layers distillation learning rate $5e^{-5}$ and prediction layer distillation learning rate $3e^{-5}$.
We performed several exploratory studies where we changed the learning parameters of distillation. 

This preliminary investigation show that changing parameters at intermediate layers distillation yields no reductions in $\sigma(X^{train})$ values. To no avail, we varied epochs \{3, 6, 9\}, learning rate \{$5e^{-7}$, $1e^{-6}$, $5e^{-5}$, $1e^{-4}$, $3e^{-4}$, $5e^{-3}$\}, batch size \{28, 32, 36\}, and weight decay \{$1e^{-4}$, $1e^{-3}$, $1e^{-2}$\}.

We experimented with the hyperparameters at prediction layers distillation in the following ranges: epochs \{2, 3, 4, 5, 6\}, learning rate \{$3e^{-6}$, $1e^{-5}$, $3e^{-5}$, $7e^{-5}$, $4e^{-4}$, $5e^{-4}$, $8e^{-4}$\}, batch size \{28, 32, 34, 36, 38, 40\}, and weight decay \{$1e^{-4}$, $1e^{-3}$, $5e^{-3}$, $1e^{-2}$, $5e^{-2}$\}. Changing only the parameters of the prediction layers distillation reduced the values of $\sigma(X^{train})$ so that $\Tilde{S}$ models for MRPC and CoLA hold $\varphi_{cnf}$. However, in order for the RTE task to satisfy $\varphi_{cnf}$, it requires the hyperparameter tuning for both prediction and intermediate layer distillations. 

The hyperparameters of the original knowledge distillation
($lr_{stg1}$, $lr_{stg2}$, $batch_{stg2}$, $epoch_{stg2}$, $wd_{stg2})$ 
are equal to $(5e^{-5}, 3e^{-5}, 32, 3, 1e^{-4})$, where ``stg1'' and ``stg2'' correspond to intermediate and prediction layer distillation respectively, ``lr'' and ``wd'' denote learning rate and weight decay. The resulting property improvement is achieved using the fine-tuning with the following parameters on RTE: ($\bm{1e^{-4}}$, $3e^{-5}$, $
\bm{36}$, $\bm{4}$, $1e^{-4})$ for $\Tilde{S}_{6L}$ and ($5e^{-5}$, $3e^{-5}$, $32$, $\bm{4}$, $1e^{-4}$) for $\Tilde{S}_{4L}$; MRPC: ($5e^{-5}$, $\bm{1e^{-5}}$, $32$, $\bm{4}$, $1e^{-4}$) for $\Tilde{S}_{6L}$ and ($5e^{-5}$, $3e^{-5}$, $32$, $\bm{2}$, $1e^{-4}$) for $\Tilde{S}_{4L}$ and CoLA: ($5e^{-5}$, $\bm{1e^{-5}}$, $32$, $\bm{4}$, $1e^{-4}$) for $\Tilde{S}_{6L}$ and ($5e^{-5}$, $\bm{1e^{-5}}$, $32$, $\bm{4}$, $1e^{-4}$) for $\Tilde{S}_{4L}$.

Models $\Tilde{S}_{6L}$ were fine-tuned for all three tasks to satisfy $\varphi_{cnf}$ property without any significant drop in the accuracy. 
We can see that 6L RTE model required the most changes to the original parameters, including the changes on the stage 1 of distillation. This can be attributed to the fact that it is the smallest training dataset. The accuracy of $\Tilde{S}_{6L}$ and $B$ on the Ev dataset are comparable, indicating that the fine-tuning did not change the regime of the distillation significantly.





To answer RQ2, we are able to fine-tune all three models to satisfy $\varphi_{cnf}$ property without the changes of the distillation architecture and training sets while avoiding significant drop in accuracy.

\subsection{Discussion} 
Based on the information presented, we can conclude that $\varphi_{cnf}$ is a non-trivial property, and satisfying the distillation quality condition guarantees its preservation. 


The architecture of the distilled model, such as the number of transformer layers as well as learning hyperparameters at the prediction layer greatly impact $\varphi_{cnf}$, while an early intermediate distillation layer has a much lesser effect on $\sigma(X^{train})$ values.

Monitoring $\varphi_{cnf}$ property therefore guides distillation hyperparameters fine-tuning, so that both accuracy and confidence are preserved. We note that this does not require any architectural modifications to the existing distillation models.

\section{Related Work} \label{sec:related}
~~~~\textbf{Black-box Equivalency Checking:}
Similar to our work, black-box equivalence checking determines functional equivalence of two programs using concrete inputs. This approach is used in verifying correctness of code compilation by establishing relations between the original and compiled programs~\cite{Kurhe:CGO22,Dahiya:APLAS17,Lim:CGO22}. After running tests, the technique discovers possible relationship between variables of two versions of the program. Next, using formal methods, it verifies that the identified equivalence relation indeed holds for all program inputs. Our approach only establishes the relationship between $S$ and $B$ confidence values, and does not generalize it to all possible inputs to those models.


\textbf{Abstraction for Large-scale Deep Models:}
The latest research in the area of formal methods for AI does not perform the verification on the state of the art large language models. There are two papers exploring relatively large deep learning models with more than ten million parameters, where one relies on formula-based approach~\cite{huang2017safety} and another on the abstraction via state transition system~\cite{du2019deepstellar}. The main challenge to make formal methods scalable to the sizes of realistic deep neural networks is the prohibitive size and complexity of a formula in these approaches~\cite{katz2017reluplex}. Implicit abstraction on the other hand does not require exact formula and state transition representation, hence it is a natural fit for verification of large models.
Several papers addressed the verification of deep neural networks by the development of abstractions. 
Idealized real-valued abstraction was proposed in \cite{henzinger2021scalable} to verify relatively small visual deep neural networks. The idea is to quantize all the operations in the network to 32-bits, and then feed into SMT solver. In the recent work~\cite{guidotti2020verification}, the authors present verification of feed forward neural network using training pipeline based on the pruning with the goal to make the model amenable to formal analysis. However, the network under verification has only fully connected layers that were reduced using pruning for network reduction and then processing by an SMT solver. 


\textbf{Distillation and Property Preservation:}  
There is a substantial research on measuring and calibrating confidence, including techniques such as temperature scaling~\cite{guo2017calibration} and on/off manifold regularization~\cite{kong2020calibrated}. However, to the best of our knowledge, the work on confidence property analysis for the knowledge distillation is limited. 
Other researchers also note that preserving properties during distillation is important, specifically the robustness property preservation. In~\cite{goldblum2020adversarially}, the method of the Adversarial Robust Distillation (ARD) was proposed to mitigate the cases when the robustness of a distilled model suffers compared to the robustness of the original model.  
However, the focus of our paper is the confidence property. Therefore, we did not use robust variants of the distillation such as ARD or Defensive Distillation in order to fairly evaluate distilled abstraction of the large-scale language models.



\section{Conclusion and Future Work} \label{sec:conclusion}

In this work, we study the confidence preservation property under a knowledge distillation abstraction. To the best of our knowledge, this is the first work to view knowledge distillation as an abstraction, as well as to study property preservation under knowledge distillation and defining the pairwise confidence preservation criterion.

We evaluate $\sigma(X^{train})$ using a black-box equivalence approach on six tasks from the linguistic benchmark dataset. For the tasks where the property fails, we modified the hyperparameters of the distillation process to ensure preservation. Our evaluations demonstrate that preservation of this confidence property could aid the distillation process by guiding the selection of hyperparameters.

Our future work will involve using formal methods to study more properties of large deep neural networks. Additionally, due to the complexity of the distillation schema for TinyBERT, not all the learning parameters were examined. Therefore, further research is needed to investigate the impact of pre-trained distillation and the combination of distillation losses.

\bibliographystyle{splncs04}
\bibliography{reference}





\end{document}